\documentclass[10pt,twocolumn,letterpaper]{article}

\usepackage[pagenumbers]{cvpr} 

\usepackage[ruled,vlined]{algorithm2e}
\usepackage{multirow}
\usepackage{tcolorbox}

\definecolor{cvprblue}{rgb}{0.21,0.49,0.74}
\usepackage[pagebackref,breaklinks,colorlinks,allcolors=cvprblue]{hyperref}

\def\paperID{*****} 
\def\confName{CVPR}
\def\confYear{2026}

\title{BlendFusion - Scalable Synthetic Data Generation for Diffusion Model Training}

\author{Thejas Venkatesh\thanks{Equal contribution, work done at Stanford University}\\
Samaya AI, Inc.\\
{\tt\small thejas@stanford.edu}
\and
Suguna Varshini Velury\footnotemark[1]\\
Independent Researcher\\
{\tt\small sugunav@stanford.edu}
}

\begin{document}
\maketitle

\begin{abstract}

With the rapid adoption of diffusion models, synthetic data generation has emerged as a promising approach for addressing the growing demand for large-scale image datasets. However, images generated purely by diffusion models often exhibit visual inconsistencies, and training models on such data can create an autophagous feedback loop that leads to model collapse, commonly referred to as Model Autophagy Disorder (MAD). To address these challenges, we propose BlendFusion\footnote{\href{https://github.com/VThejas/blendfusion}{https://github.com/VThejas/blendfusion}}, a scalable framework for synthetic data generation from 3D scenes using path tracing. Our pipeline incorporates an object-centric camera placement strategy, robust filtering mechanisms, and automatic captioning to produce high-quality image–caption pairs. Using this pipeline, we curate FineBLEND\footnote{\href{https://huggingface.co/datasets/vThejas/FineBLEND}{https://huggingface.co/datasets/vThejas/FineBLEND}}, an image-caption dataset constructed from a diverse set of 3D scenes. We empirically analyze the quality of FineBLEND and compare it to several widely used image–caption datasets. We also demonstrate the effectiveness of our object-centric camera placement strategy relative to object-agnostic sampling approaches. Our open-source framework is designed for high configurability, enabling the community to create their own datasets from 3D scenes.
\end{abstract}    
\section{Introduction}
\label{sec:intro}

Recent advances in diffusion models have significantly improved the quality of generative image synthesis. Modern text-to-image models such as Stable Diffusion \cite{rombach2022highresolutionimagesynthesislatent} and Imagen \cite{saharia2022photorealistictexttoimagediffusionmodels} are trained on large-scale image–caption datasets, where textual descriptions guide the generation process. The success of these models largely depends on the availability of high-quality paired image–text datasets. Large-scale datasets such as LAION-5B \cite{schuhmann2022laion5bopenlargescaledataset} and MS-COCO \cite{lin2015microsoftcococommonobjects} contain billions of image–text pairs and have enabled the training of such models. 

However, collecting and curating such datasets at scale presents several challenges, including annotation cost \cite{lin2015microsoftcococommonobjects}, dataset bias \cite{garcia2023uncuratedimagetextdatasetsshedding}, and privacy and safety concerns \cite{relaion}. As a result, synthetic data generation has emerged as a promising approach for augmenting or replacing real-world training data \cite{electronics13173509}, \cite{gupta2021transitioningrealsyntheticdata}.

Among synthetic data approaches, diffusion-based image generation has gained significant popularity due to its ability to produce visually compelling images at scale \cite{bertazzini2025dragonlargescaledatasetrealistic}, \cite{saragih2024usingdiffusionmodelsgenerate}.

Despite their impressive capabilities, diffusion-generated images often suffer from visual inconsistencies such as geometric distortions, object hallucinations, and semantic artifacts \cite{bertazzini2025dragonlargescaledatasetrealistic}, \cite{ouyang2025consistencycriticcorrectinginconsistencies}. These issues limit their usefulness as reliable training data.

Furthermore, recent studies have shown that training generative models on synthetic data produced by other generative models can lead to an autophagous feedback loop, where the model gradually degrades in quality. This phenomenon, known as Model Autophagy Disorder (MAD), can ultimately lead to model collapse \cite{alemohammad2023selfconsuminggenerativemodelsmad}, \cite{shumailov2024curserecursiontraininggenerated}.

An alternative approach is to generate synthetic data using 3D scenes and physically-based rendering \cite{richter2016playingdatagroundtruth}, \cite{7780721}. Compared to diffusion-generated images, path-traced images provide physically consistent geometry, lighting, and object structure. Such data generation pipelines offer fine-grained control over scene selection, camera placement, and metadata collection, making them attractive for producing high-quality training datasets \cite{richter2016playingdatagroundtruth}, \cite{tremblay2018fallingthingssyntheticdataset}. 

Moreover, large online repositories of user-contributed 3D assets and environments, such as BlenderKit\footnote{\url{https://www.blenderkit.com}} and Sketchfab\footnote{\url{https://sketchfab.com}}, contain several downloadable assets and environments, many released under permissive licenses, enabling scalable collection of diverse scenes for synthetic rendering pipelines.

\begin{figure*}[t]
    \centering
    \includegraphics[width=\textwidth]{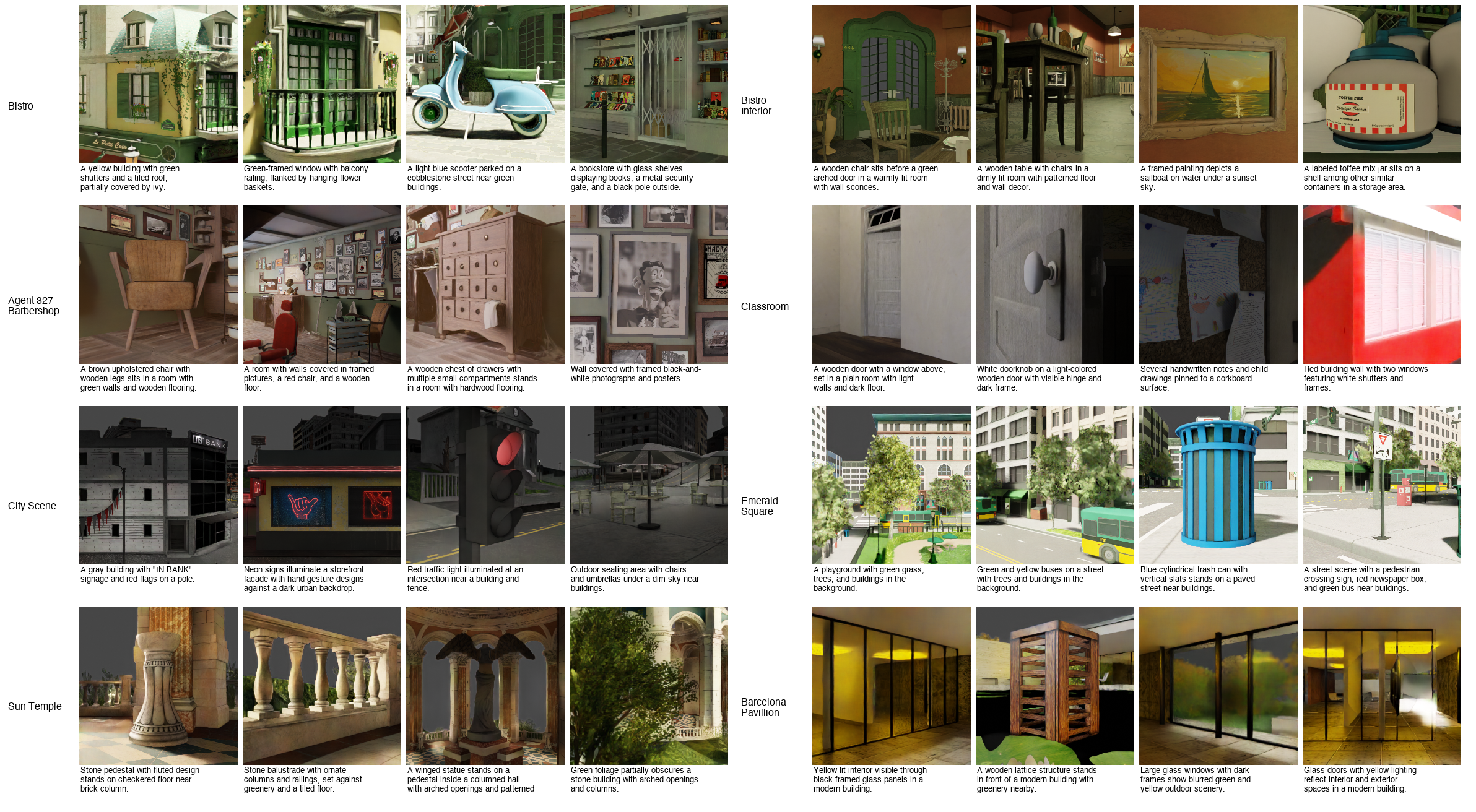}
    \caption{Images generated by BlendFusion for different scenes}
    \label{fig:blendfusion_images}
\end{figure*}

In this work, we propose BlendFusion, a scalable framework for generating high-quality synthetic image–caption pairs from 3D scenes using path tracing. BlendFusion incorporates a novel object-centric camera placement strategy that enhances the discovery of `useful' camera placements. We further employ heuristics and vision language model  based filtering to remove visually subpar renderings. Upon filtering, the final dataset is curated by eliminating similar images through a diversity-aware sampling strategy.

Our main contributions are as follows:
\begin{itemize}
\item We introduce BlendFusion, a pipeline for generating image–caption pairs from 3D scenes using object-centric rendering and automated filtering.
    \item Using this pipeline, we construct FineBLEND, a dataset of 7,500 rendered image–caption pairs derived from diverse 3D environments.
    \item We analyze the resulting dataset using image-caption alignment metrics and aesthetic metrics and compare it to several widely used image–caption datasets.
    \item We demonstrate that object-centric camera placement improves the yield of usable renders compared to object-agnostic sampling strategies.
\end{itemize}

\section{Related Work}

\begin{figure*}[t]
    \centering
    \includegraphics[width=\textwidth]{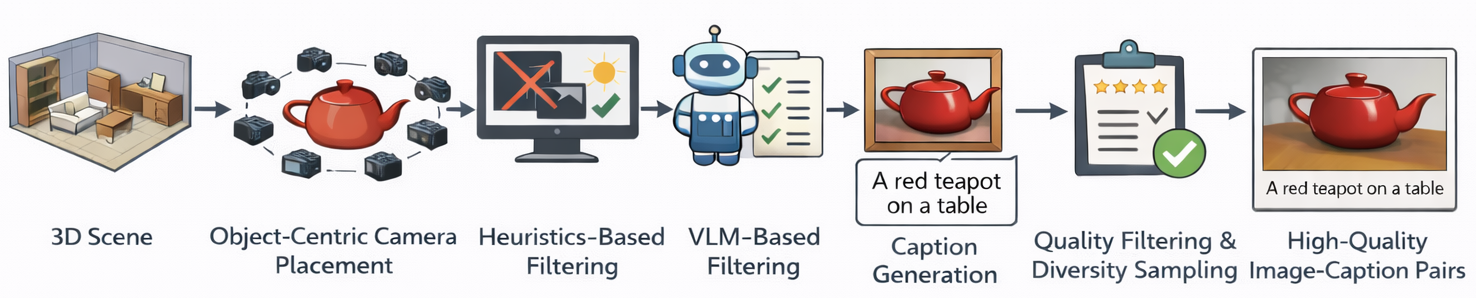}
    \caption{The BlendFusion Pipeline}
    \label{fig:blendfusion_pipeline}
\end{figure*}

\paragraph{Diffusion models and synthetic training data.}
Diffusion models have emerged as the dominant paradigm for high-fidelity image synthesis, achieving state-of-the-art results in text-to-image generation and related visual generation tasks \cite{ho2020denoisingdiffusionprobabilisticmodels, nichol2021improveddenoisingdiffusionprobabilistic, rombach2022highresolutionimagesynthesislatent}. Their success has significantly increased the demand for large-scale, diverse image datasets. Synthetic data generation using diffusion models has therefore become an attractive strategy for scaling training corpora and augmenting existing datasets \cite{wu2023datasetdmsynthesizingdataperception, wang2023diffusiondblargescalepromptgallery}. For example, recent work such as \textsc{DRAGON} constructs large-scale collections of synthetic images generated by multiple diffusion models, producing millions of samples to support tasks such as synthetic image detection and benchmarking \cite{bertazzini2025dragonlargescaledatasetrealistic}. Such datasets illustrate the growing role of generative models in dataset construction.

\paragraph{Self-training and model collapse.}
Despite the scalability of diffusion-based synthesis, training models on their own generated outputs can introduce instability. Recursive training on synthetic samples leads to a progressive degradation in either fidelity or diversity, a phenomenon termed \emph{Model Autophagy Disorder} (MAD) \cite{alemohammad2023selfconsuminggenerativemodelsmad}. In these autophagous training loops, artifacts and biases accumulate across generations unless sufficiently large amounts of fresh real data are injected \cite{alemohammad2023selfconsuminggenerativemodelsmad}. Recent work further investigates whether such self-training can be leveraged constructively. For instance, \textsc{Neon} proposes a learning strategy that uses the degradation signal from self-training to improve generative models via negative extrapolation, mitigating the collapse that typically arises in naive synthetic self-training \cite{alemohammad2025neonnegativeextrapolationselftraining}. These studies highlight both the promise and the risks of relying on model-generated data.

\paragraph{Synthetic data from 3D scenes.}
An alternative approach is to generate training data from 3D assets and scenes using physically based rendering. Graphics-driven pipelines such as BlenderProc \cite{denninger2019blenderproc} and Kubric \cite{greff2022kubricscalabledatasetgenerator} enable scalable simulation, rendering, and annotation of synthetic visual data. Large repositories of 3D assets, including Objaverse \cite{deitke2022objaverseuniverseannotated3d} and Objaverse-XL \cite{ deitke2023objaversexluniverse10m3d}, further enable dataset construction at unprecedented scale, containing millions of diverse 3D objects collected from the web.

\paragraph{Captioning rendered 3D content.}
Several recent works investigate generating language supervision from rendered views of 3D assets. Cap3D \cite{luo2023scalable3dcaptioningpretrained} constructs large-scale 3D–text datasets by rendering objects from multiple viewpoints and generating captions using pretrained vision–language models. However, caption quality is highly sensitive to the chosen viewpoint; atypical renderings can lead to hallucinated or inaccurate descriptions. DiffuRank \cite{luo2025viewselection3dcaptioning} addresses this issue by ranking candidate views based on their alignment with the underlying 3D object before caption generation, improving caption accuracy and dataset quality. 
\section{BlendFusion}

We propose \textbf{BlendFusion}, an automatic pipeline for generating high-quality image-caption pairs from 3D scenes. BlendFusion uses the Blender\footnote{\url{https://www.blender.org/}} graphics engine to render physically accurate images through path tracing. The pipeline is developed using the BlenderProc library \cite{denninger2019blenderproc}, which simplifies large-scale  procedural rendering in Blender. Figure \ref{fig:blendfusion_pipeline} describes the end-to-end pipeline. 

The following sections discuss each stage of the pipeline in detail.

\subsection{Object-centric Camera Placement}
\label{sec:cam_placement}

In order to generate meaningful images, picking the right subject for the image and ensuring proper framing is crucial \cite{luo2025viewselection3dcaptioning}. Several state-of-the-art datasets for training text-to-image models use common objects as the image's central subject \cite{lin2015microsoftcococommonobjects}, \cite{5206848}. 

This motivates us to adopt an object-centric camera placement strategy designed to systematically capture diverse yet controlled views of individual objects within a 3D scene.  Rather than sampling cameras globally at random, our method places cameras relative to each mesh object, ensuring that the rendered image set provides consistent coverage, predictable framing, and interpretable metadata for downstream filtering.

For each mesh object, we retrieve the object's axis-aligned bounding box and calculate its center. To generate candidate viewpoints, we place cameras on a discrete orbit around the object center. Azimuth angles are sampled uniformly every $45^\circ$, yielding eight viewpoints around the object, while elevation is fixed at $0^\circ$. We fix the elevation to 0° to approximate a natural eye-level viewpoint. Thus, all cameras lie on a horizontal ring centered on the object. For each azimuth $\phi$ and elevation $\theta$, a unit viewing direction
\[
v(\phi,\theta)=
\begin{bmatrix}
\cos\theta\cos\phi \\
\cos\theta\sin\phi \\
\sin\theta
\end{bmatrix}
\]
is constructed and normalized. The camera is then oriented with a look-at transform so that it always points toward the bounding box's center.

By sampling several azimuths around the object, we ensure that different perspectives of the object are captured. Also, in case any specific camera placement is not usable due to occlusions, the object is still likely captured by another placement.

To ensure optimal framing and to maintain a consistent object scale across rendered views, we adapt the camera distance based on the projected size of the object. For each viewpoint, the object's bounding box is projected onto the plane perpendicular to the viewing direction. Let $h_o$ denote half of the projected vertical span of this bounding box. Given a target fill fraction $f$ (the ratio of the projected vertical span to the height of the image) and vertical field of view $\mathrm{FOV}_y$, the camera distance is computed as
\[
d = \frac{h_o}{f \tan(\mathrm{FOV}_y/2)}.
\]
This formulation ensures that the object occupies an approximately constant fraction of the image regardless of its physical size or orientation relative to the camera. We choose $f = 2/3$, i.e. the object occupies $\frac{2}{3}^{rd}$  of the image vertically, around the center of the image.

Based on the chosen camera distance and the 8 azimuths, 8 images are rendered per object. We choose a resolution of $256 \times 256$ to balance efficiency and image quality. 

The pipeline is implemented such that the desired azimuths, elevations, fill fractions and image resolution are configurable. Furthermore, we implement a minimum bounding box diagonal filter to support filtering small objects, however we do not apply this for our experiments. 

The entire algorithm is shown in Algorithm~\ref{alg:camera}.

\begin{algorithm}[t]
\caption{Object-Centric Camera Placement}
\label{alg:camera}

\KwIn{Object bounding box $B_o$, camera FOV $\mathrm{FOV}_y$, fill fraction $f$, azimuth set $\mathcal{A}$, elevation set $\mathcal{E}$}
\KwOut{Camera poses $\mathcal{P}$}

Compute object center $c_o$ from $B_o$\;
$\mathcal{P} \leftarrow \emptyset$\;

\For{$\theta \in \mathcal{E}$}{
    \For{$\phi \in \mathcal{A}$}{
        $v \leftarrow [\cos\theta\cos\phi,\; \cos\theta\sin\phi,\; \sin\theta]$\;
        Normalize $v$\;

        Project $B_o$ onto plane orthogonal to $v$\;
        Compute projected half-height $h_o$\;

        $d \leftarrow \frac{h_o}{f \tan(\mathrm{FOV}_y/2)}$\;
        $p \leftarrow c_o + d\,v$\;

        $T \leftarrow \textsc{LookAt}(p, c_o)$\;
        $\mathcal{P} \leftarrow \mathcal{P} \cup \{T\}$\;
    }
}

\Return $\mathcal{P}$

\end{algorithm}

\subsection{Heuristics-Based Filtering}
\label{sec:heuristic_filter}

Although our object-centric camera placement produces consistent viewpoints, some rendered images can still be unsuitable for training due to lighting artifacts, empty frames, or low-information content. To remove such cases, we apply a lightweight set of heuristic filters based on statistics computed from the rendered RGB images and segmentation maps.

First, we discard images where the target object occupies no pixels in the segmentation map (i.e., zero fill). These cases typically arise when the object falls outside the camera frustum or becomes fully occluded by other scene geometry. Removing such frames prevents training on images that contain no visual signal for the intended object.

Next, we filter images based on brightness. Extremely dark images often result from poor lighting configurations or objects located in shadowed regions of the scene. We compute the mean brightness of the image (after converting to grayscale) and discard images whose average brightness falls below a desired threshold. This removes frames where the object is barely visible.

We also remove images with very low pixel variance. Low variance indicates that the image contains little visual structure, which may occur in cases where the frame is dominated by flat backgrounds or uniformly colored surfaces. We compute the grayscale pixel variance and discard images which are below a specified threshold.

Finally, we filter images that are predominantly dark. We compute the fraction of pixels whose intensity falls below a small threshold ($brightness <= 5$) and discard images where the fraction of near-black pixels  is greater than a certain threshold. This criterion further removes underexposed renders and frames where the scene lighting fails to illuminate the object adequately.

Together, these simple heuristics provide an efficient first-pass quality filter that eliminates degenerate renders while preserving the majority of visually informative images for dataset construction.

\subsection{VLM-Based Filtering}
\label{sec:vlm_filter}

In addition to heuristic filtering, we apply a vision--language model (VLM) to remove images that are difficult to caption reliably. We use \textit{Qwen3-VL-8B-Instruct} \cite{qwen3technicalreport} to evaluate the captionability of each rendered image.

Images are accepted if they satisfy at least one of two criteria: (1) an \emph{object-centric} criterion, where a recognizable object or character is visible with sufficient context to identify it, or (2) a \emph{scene-centric} criterion, where the overall environment is clearly identifiable (e.g., forest, street, indoor room). 

The model is instructed to reject common failure modes including extreme close-ups, partial fragments lacking sufficient context, severe object truncation, or images where the object or scene category cannot be determined. Low resolution alone is not considered grounds for rejection; however, the model is prompted to be strict about rejecting images that can only be vaguely captioned. The entire prompt can be found in Appendix~\ref{app:prompts}.

This VLM-based filtering provides a  concrete semantic quality check in addition to the earlier heuristic filters, by ensuring that retained images only contain identifiable objects or scenes suitable for descriptive captions.

\subsection{Caption Generation}
\label{sec:captions}

This stage of the pipeline generates captions for the filtered images. For each remaining image, we prompt a vision--language model (\textit{Qwen3-VL-8B-Instruct}) to produce a single factual caption describing the visible content. The prompt instructs the model to describe only what is clearly observable in the image and to avoid hallucinating unseen objects or attributes. Captions are required to be short, neutral, and concrete. The entire prompt can be found in Appendix~\ref{app:prompts}.

\subsection{Quality-Based Filtering}
\label{sec:quality_filter}

As an optional step, we apply an additional quality filtering stage based on image–text alignment and perceptual quality scores. Specifically, we compute CLIPScore \cite{hessel2022clipscorereferencefreeevaluationmetric} and LAION aesthetic predictor score \cite{schuhmann2022laion_aesthetics} for each rendered image and discard samples whose scores fall below predefined thresholds.

The CLIPScore measures the semantic alignment between an image and its associated caption using a pretrained CLIP model \cite{radford2021learningtransferablevisualmodels}. Low CLIP scores typically indicate mismatches between the visual content and the generated caption or images whose content is difficult to describe reliably. The LAION aesthetic predictor score, predicted using a lightweight aesthetic predictor trained on human preference data, estimates the perceptual quality of the image. Low aesthetic scores often correspond to visually unappealing renders, such as poorly lit scenes, low-detail surfaces, or images dominated by artifacts.

Images with CLIPScore or aesthetic score below the specified thresholds are removed prior to the subsequent dataset construction step. This filtering stage helps ensure that the remaining images exhibit both strong semantic alignment with their captions and acceptable perceptual quality.

\subsection{Diversity-Aware Sampling}
\label{sec:diversity}

After filtering, we perform diversity-aware sampling to select a subset of images that maximizes visual coverage while minimizing redundancy. Each image is embedded using DinoV2 \cite{oquab2024dinov2learningrobustvisual} and the embeddings are $\ell_2$-normalized. Pairwise distances are computed in the embedding space, and we apply \emph{farthest point sampling} (FPS) \cite{articlefps}, which iteratively selects the image that maximizes its minimum distance to the set of already selected samples. This procedure encourages the selected subset to span the visual distribution of the rendered data while avoiding near-duplicate views.

When constructing train, test, and validation splits, we extend FPS to operate across multiple splits simultaneously. The procedure begins by selecting an initial set of diverse seed images using standard FPS, which are then assigned across the splits to initialize each set. During subsequent iterations, selections are made from the remaining pool of images. We perform a weighted round-robin assignment across dataset splits, based on the split ratio. For each split, the algorithm selects the image that is farthest (in embedding space) from the images already assigned to that split. This process continues until the desired number of samples has been allocated to each split.

By maintaining separate distance criteria for each split while sampling from a shared pool of images, this parallel FPS strategy ensures that each split remains internally diverse while avoiding redundancy across the selected dataset.
\section{The FineBLEND Dataset}

\begin{table*}[h]
\centering
\footnotesize
\setlength{\tabcolsep}{5pt}
\begin{tabular}{l c c c c c}
\hline
Scene & Total & Heuristic Passed & VLM Passed & CLIP Score & LAION Aesthetic Score \\
\hline
Bistro Exterior & 10368 & 4666 (45.0\%) & 3308 (31.9\%) & $24.80 \pm 3.79$ & $4.24 \pm 0.74$ \\
Bistro Interior & 9488 & 3172 (33.4\%) & 1557 (16.4\%) & $26.20 \pm 3.43$ & $4.27 \pm 0.68$ \\
Pavillion & 840 & 329 (39.2\%) & 203 (24.2\%) & $25.27 \pm 3.36$ & $3.54 \pm 0.71$ \\
Sun Temple & 4136 & 984 (23.8\%) & 630 (15.2\%) & $23.73 \pm 2.86$ & $3.80 \pm 0.78$ \\
Emerald Square & 8240 & 5105 (62.0\%) & 4397 (53.4\%) & $26.57 \pm 3.65$ & $5.04 \pm 0.94$ \\
Classroom & 1456 & 225 (15.5\%) & 91 (6.2\%) & $28.40 \pm 2.89$ & $3.22 \pm 0.55$ \\
City Scene & 760 & 280 (36.8\%) & 211 (27.8\%) & $23.49 \pm 4.09$ & $4.15 \pm 0.71$ \\
Barbershop & 15976 & 2571 (16.1\%) & 827 (5.2\%) & $26.45 \pm 4.31$ & $4.19 \pm 0.81$ \\
\hline
\end{tabular}
\caption{Scene-wise statistics for images generated by the BlendFusion pipeline before final dataset curation.}
\label{tab:scene_stats}
\end{table*}

Using the BlendFusion pipeline, we construct \textbf{FineBLEND}, a curated image-caption dataset generated from high-quality 3D scenes. Our goal is to produce synthetic data that is both visually diverse and semantically captionable while avoiding many of the artifacts commonly found in purely generative datasets.

\paragraph{Scene Sources.}
FineBLEND is generated from a collection of scenes drawn from multiple public sources, including scenes from the NVIDIA ORCA\footnote{\url{https://developer.nvidia.com/orca}} project (Bistro and Bistro Interior \cite{ORCAAmazonBistro}, Emerald Square \cite{ORCANVIDIAEmeraldSquare} and Sun Temple \cite{OrcaUE4SunTemple}), official Blender demo scenes\footnote{\url{https://www.blender.org/download/demo-files/\#cycles}} (Agent 327 Barbershop, Barcelona Pavillion and Classroom) and the City scene\footnote{\url{https://www.blenderkit.com/get-blenderkit/a1600774-b49a-4590-82f8-f663047f5c8d/}} from BlenderKit. These scenes span a wide range of environments and visual conditions, including indoor scenes, architectural spaces, urban exteriors and natural environments. They also vary significantly in lighting conditions (e.g., daylight, partial sunlight, and indoor lighting), scene scale, and geometric complexity.

\paragraph{Data Generation and Filtering.}
We render object-centric views from these scenes using the camera placement strategy described in Section~\ref{sec:cam_placement}. The initial renders are first filtered as described in Section~\ref{sec:heuristic_filter} using brightness, variance, and dark pixel threshold of 30, 300 and 0.3 respectively. We then apply the VLM-based filtering procedure described in Section~\ref{sec:vlm_filter}, which removes images that are difficult to caption reliably. After these stages, the pipeline produces a pool of high-quality candidate images. All the filtered images are now captioned automatically as described in Section~\ref{sec:captions}. We refer to this set of \textbf{11,224} image-caption pairs as the \textbf{BlendFusion} dataset. The scene-wise statistics can be found in Table~\ref{tab:scene_stats}.

\paragraph{Quality Filtering and Subsampling.}
To further improve dataset quality and diversity, images with a CLIPScore below $20$ or an aesthetic score below $3$ are removed as described in Section~\ref{sec:quality_filter}. Finally, we apply the diversity-aware sampling and splitting strategy described in Section~\ref{sec:diversity} to reduce redundancy and ensure broad visual coverage.

The resulting \textbf{FineBLEND} dataset contains \textbf{7500} images paired with automatically generated captions. The training, validation and test splits contain \textbf{4500}, \textbf{1500} and \textbf{1500} images respectively.

\begin{figure}[t]
    \centering
    \includegraphics[width=\columnwidth]{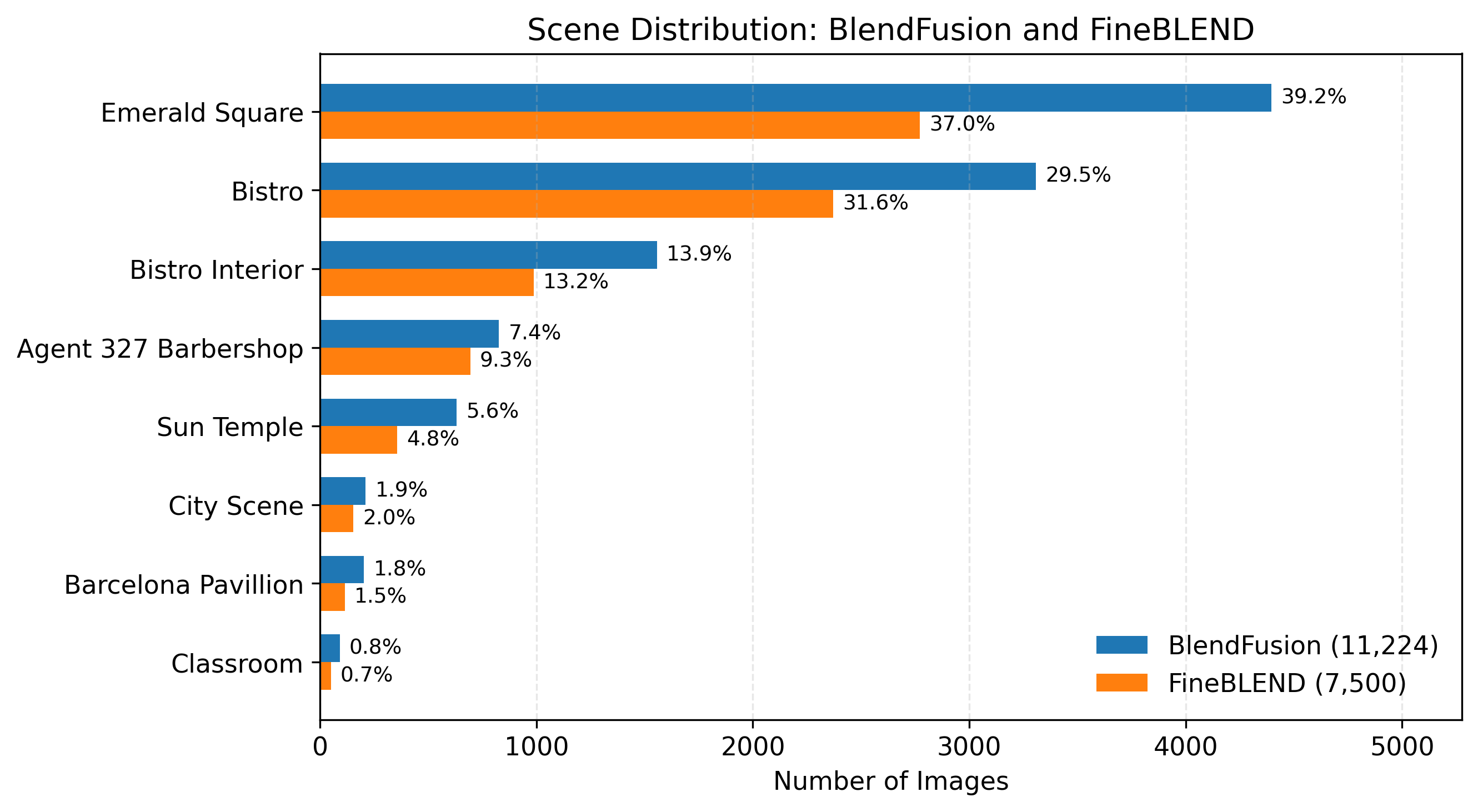}
    \caption{Scene composition for the BlendFusion and FineBLEND datasets. Bars show the number of images per scene, with percentages indicating the proportion of each scene within the respective dataset.}
    \label{fig:dataset}
\end{figure}

The scene-wise composition is shown in figure~\ref{fig:dataset}.

\section{Analysis}

\subsection{Dataset Quality}

To assess the quality of the BlendFusion and FineBLEND datasets, we compare their CLIPScore and LAION aesthetic score distributions against three widely used real-world image-caption datasets: MS-COCO~\cite{lin2015microsoftcococommonobjects}, Conceptual Captions~\cite{sharma2018conceptual}, and ReLAION~\cite{relaion}. For each reference dataset, we randomly sample 10,000 image-caption pairs and compute both scores under identical settings. Table~\ref{tab:quality} reports the results.

\begin{table}[t]
\centering
\small
\begin{tabular}{lccc}
\toprule
\textbf{Dataset} & \textbf{N} & \textbf{CLIPScore} & \textbf{Aesthetic} \\
\midrule
COCO         & 10{,}000 & 25.43 $\pm$ 3.82 & \textbf{5.79 $\pm$ 0.90} \\
CC           & 10{,}000 & 24.72 $\pm$ 4.84 & 5.22 $\pm$ 1.19 \\
ReLAION      & 10{,}000 & \textbf{28.74 $\pm$ 4.99} & 4.75 $\pm$ 1.23 \\
\midrule
BlendFusion  & 11{,}224 & 25.76 $\pm$ 3.80 & 4.50 $\pm$ 0.94 \\
FineBLEND    &  7{,}500 & 25.91 $\pm$ 3.37 & 4.52 $\pm$ 0.86 \\
\bottomrule
\end{tabular}
\caption{CLIPScore and LAION aesthetic score comparison across datasets.}
\label{tab:quality}
\end{table}

\paragraph{Image--text alignment.}
BlendFusion achieves a mean CLIPScore of 25.76, comparable to COCO (25.43) and Conceptual Captions (24.72), indicating that the automatically generated captions align well with the rendered content even before quality filtering. After filtering and diversity-aware sampling, FineBLEND's mean CLIPScore rises to 25.91 with a reduced standard deviation of 3.37, which is the lowest among all datasets. This tighter distribution reflects both the controlled rendering conditions and the effectiveness of the quality filtering stage.

Notably, the standard deviation of CLIPScores across all three reference datasets is considerably higher (3.82--4.99) than FineBLEND's (3.37). Real-world datasets are inherently noisy, web-scraped captions may be loosely associated with their images, and crowd-sourced annotations vary in specificity and accuracy. In contrast, our pipeline generates captions from a VLM that directly observes each rendered image, producing descriptions that are consistently grounded in the visual content. The low variance confirms that this controlled captioning process yields reliable image-text alignment across the dataset.

\paragraph{Perceptual quality.}
FineBLEND's mean aesthetic score (4.52) is lower than COCO (5.79) and CC (5.22), but comparable to ReLAION (4.75). We attribute this gap primarily to the domain mismatch between our rendered images and the natural photographs that the LAION aesthetic predictor was trained on. Because our images are rendered in Blender using path tracing, they exhibit characteristics that differ from natural photographs. The aesthetic predictor, having been trained on human preference judgments of photographs, systematically scores such rendered content lower even when the images are geometrically and semantically well-formed. This gap could potentially be bridged through diffusion-based post-processing which is further discussed in the conclusion section. Nonetheless, the scores are comparable to ReLAION, which is a large-scale web-scraped dataset used for training state-of-the-art models. This suggests that the rendered images are already of sufficient perceptual quality for use as training data.

\subsection{Qualitative Comparison with Diffusion-Generated Images}

\begin{figure*}[t]
\centering
\setlength{\tabcolsep}{2pt}
\begin{tabular}{ccccc}
\includegraphics[width=0.18\linewidth]{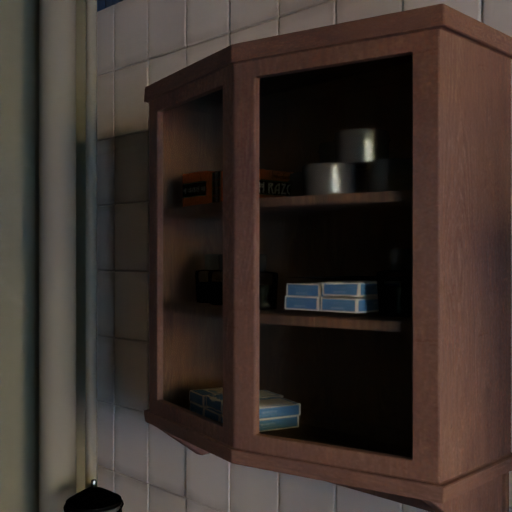} &
\includegraphics[width=0.18\linewidth]{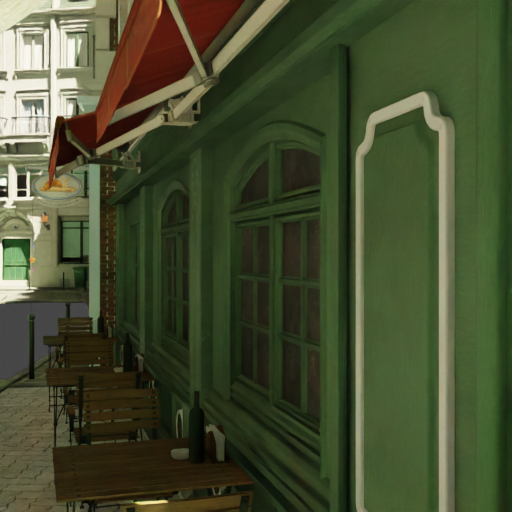} &
\includegraphics[width=0.18\linewidth]{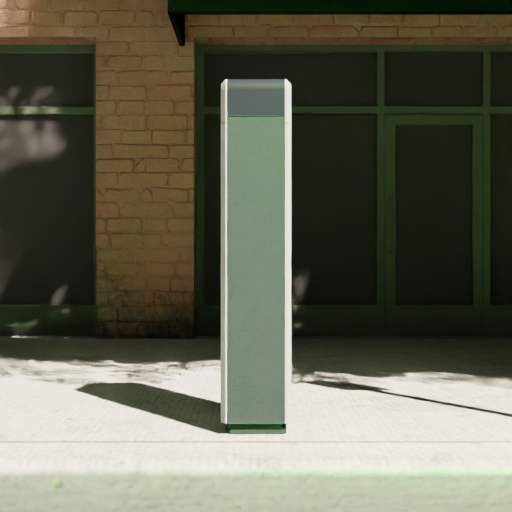} &
\includegraphics[width=0.18\linewidth]{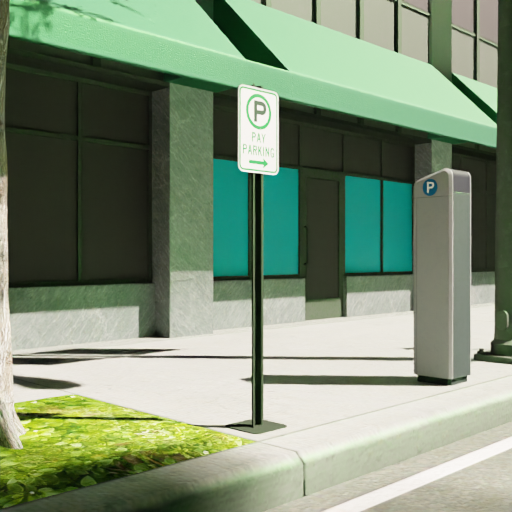} &
\includegraphics[width=0.18\linewidth]{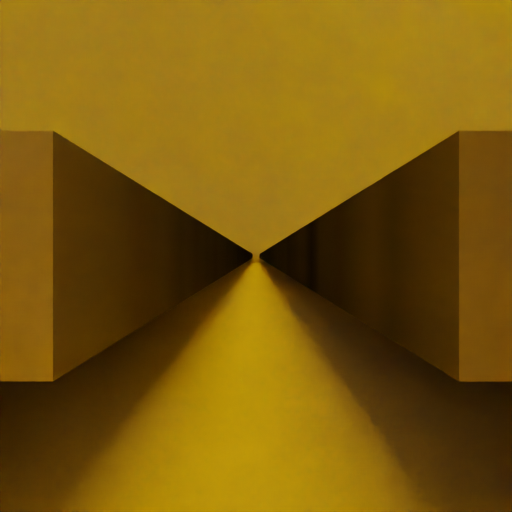} \\
\includegraphics[width=0.18\linewidth]{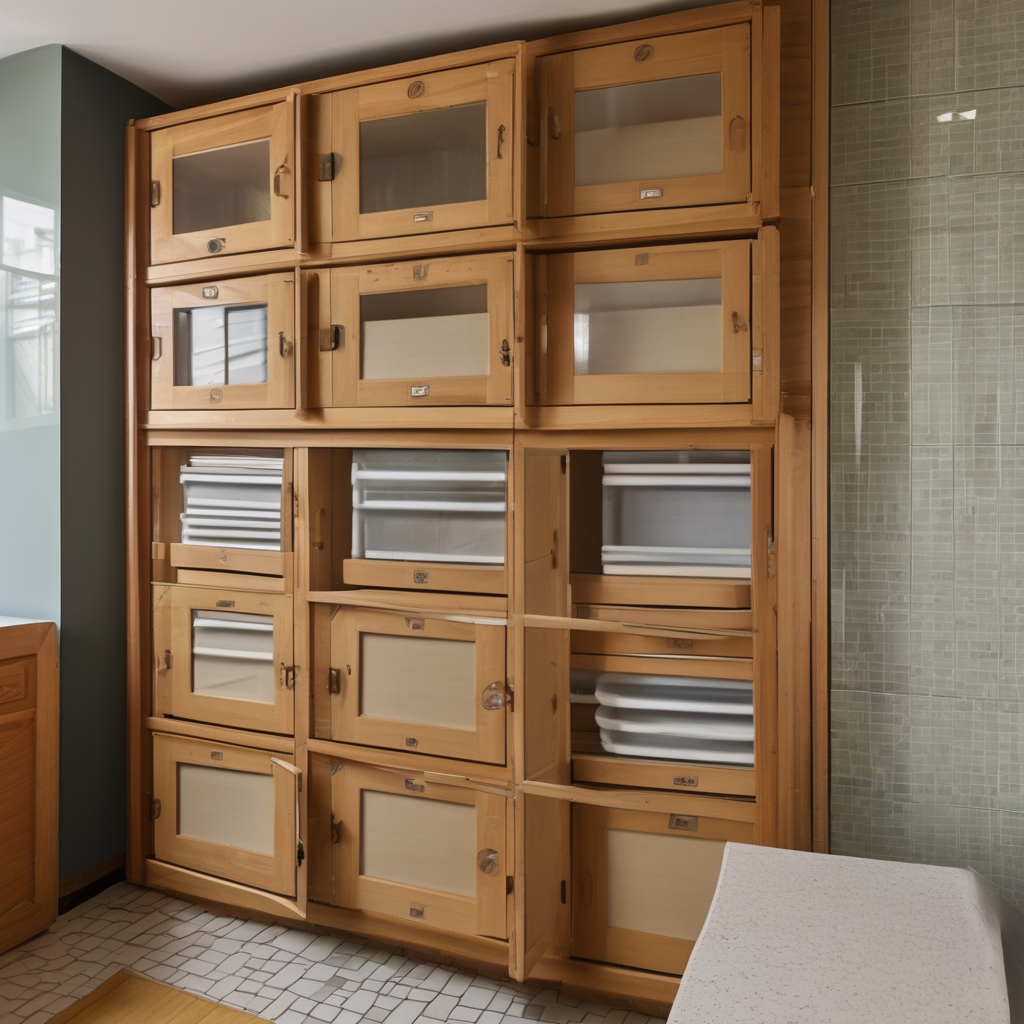} &
\includegraphics[width=0.18\linewidth]{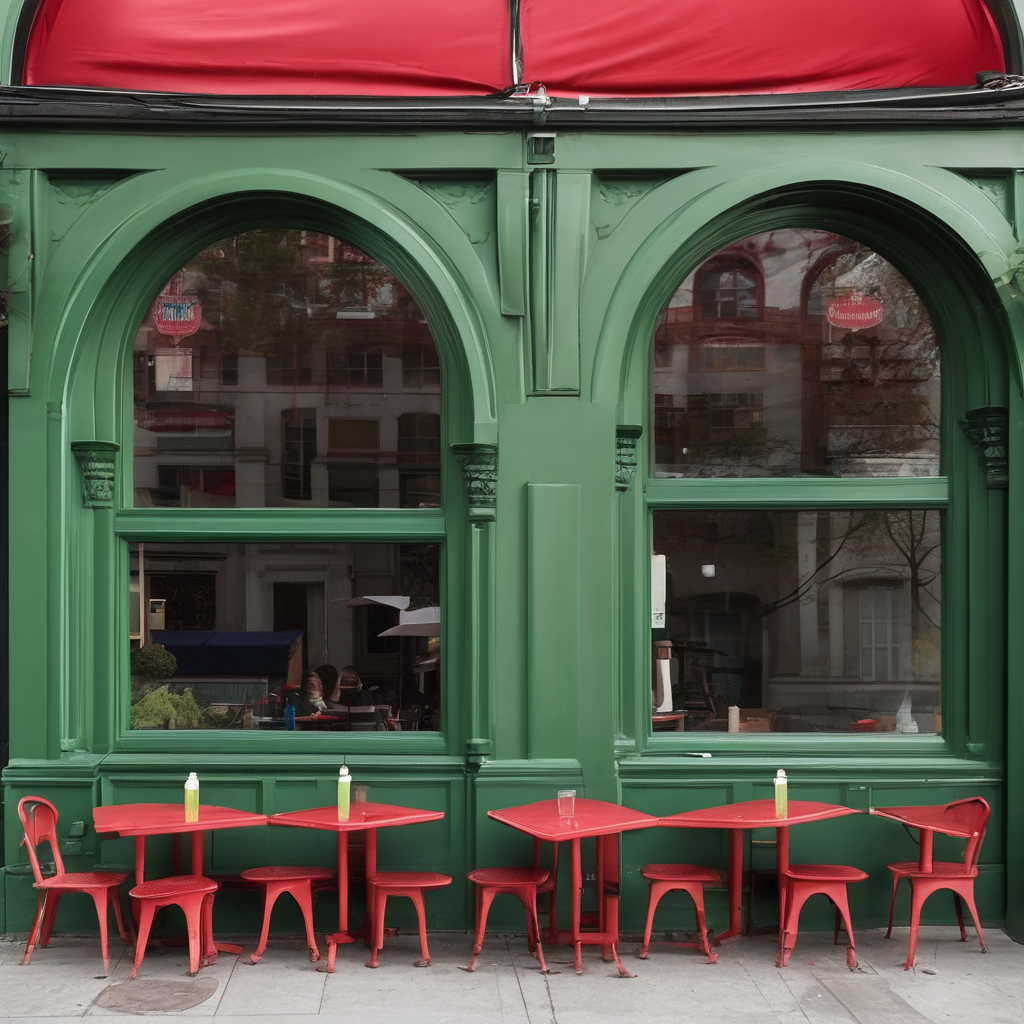} &
\includegraphics[width=0.18\linewidth]{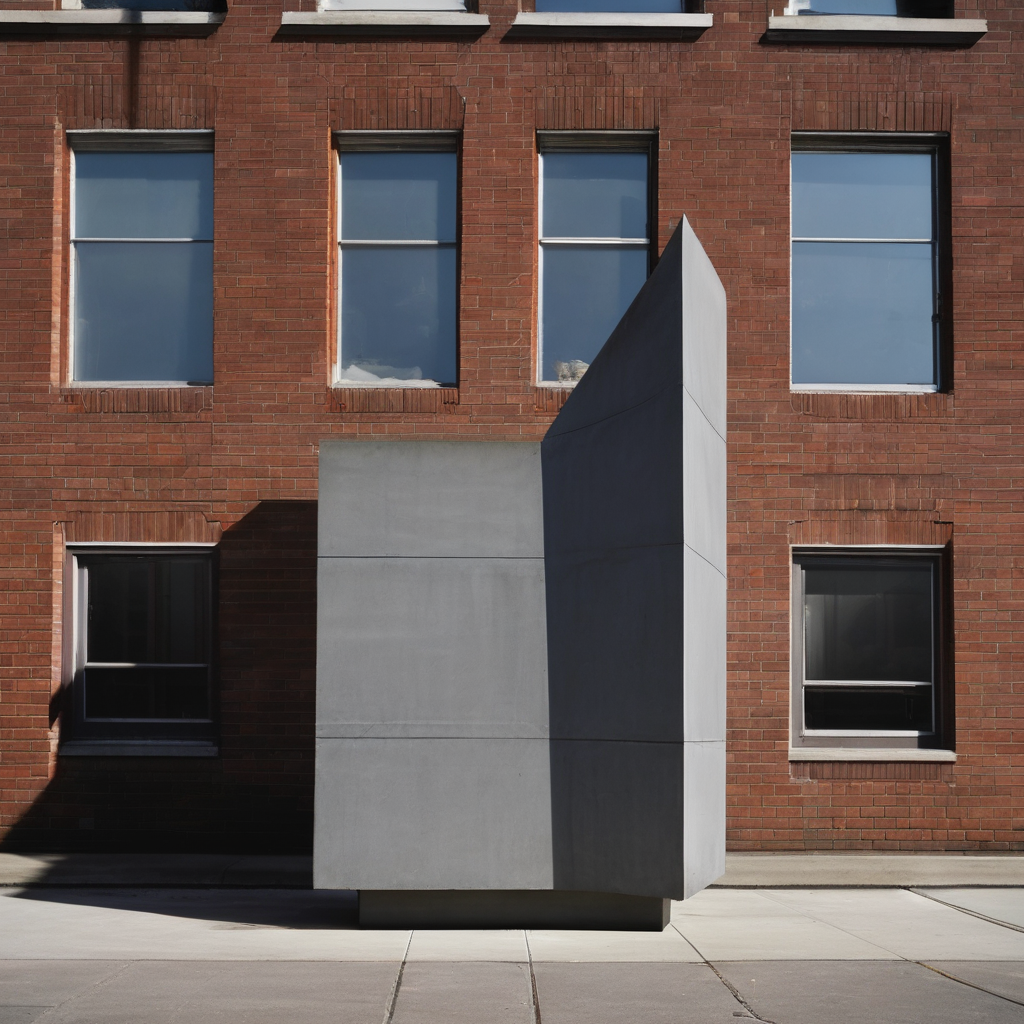} &
\includegraphics[width=0.18\linewidth]{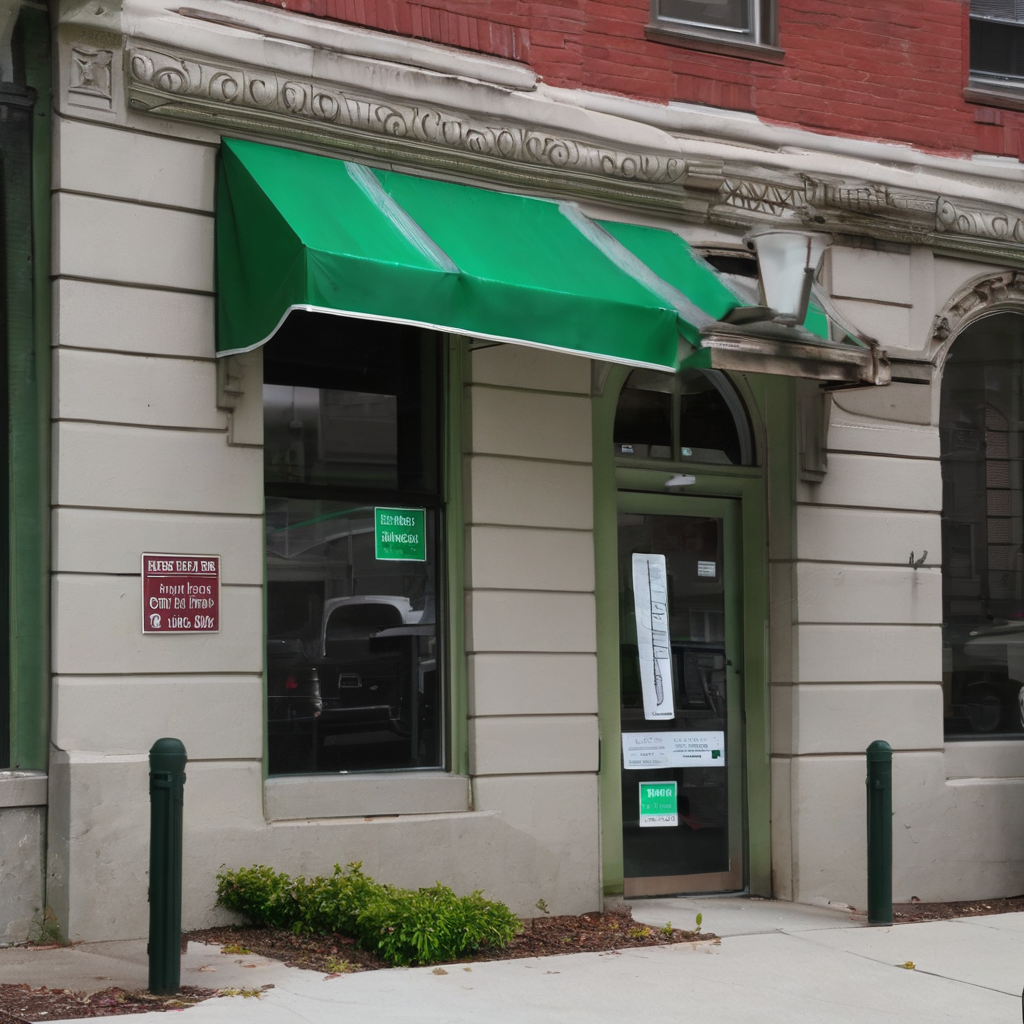} &
\includegraphics[width=0.18\linewidth]{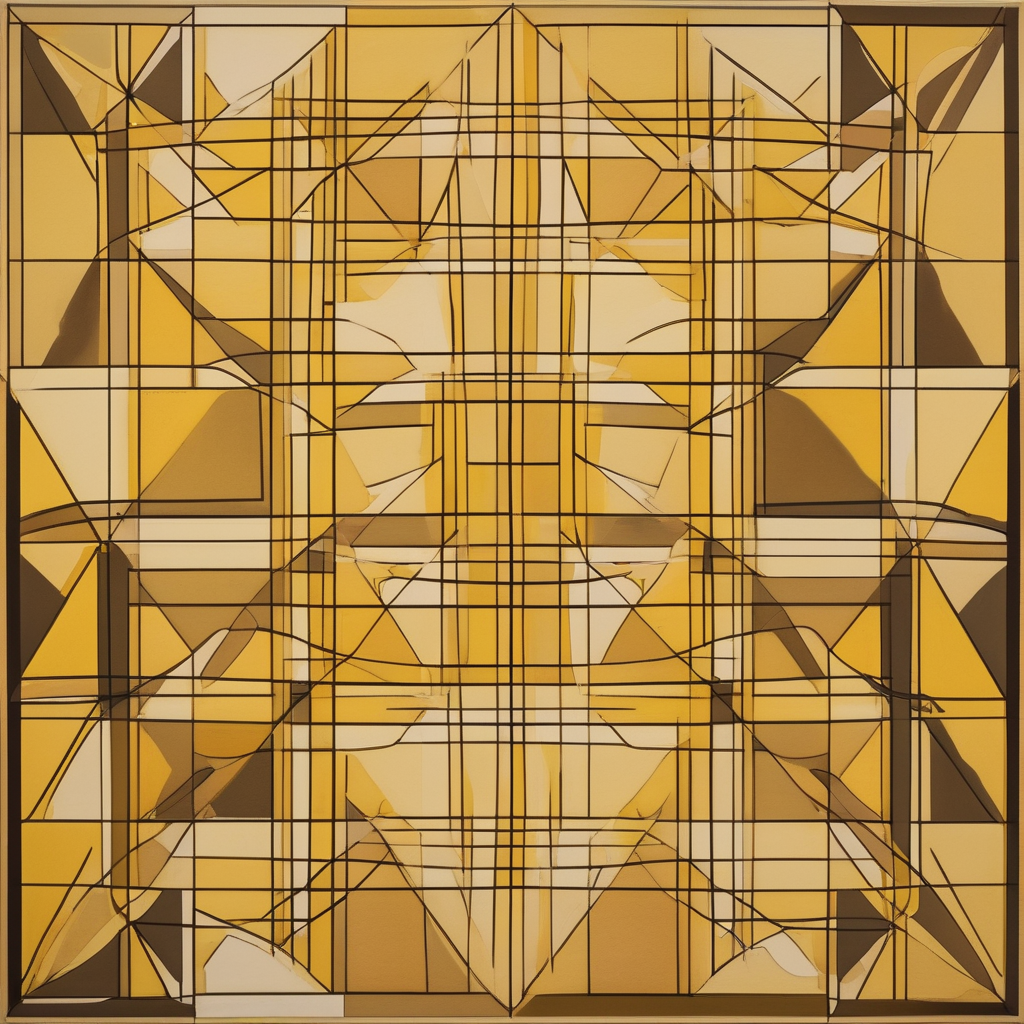} \\[2pt]
{\scriptsize \parbox{0.18\linewidth}{\centering ``Wooden cabinet with glass doors mounted on tiled wall, containing stacked boxes.''}} &
{\scriptsize \parbox{0.18\linewidth}{\centering ``Green-painted building with arched windows and outdoor tables under a red awning.''}} &
{\scriptsize \parbox{0.18\linewidth}{\centering ``A tall, gray rectangular object stands on pavement in front of a brick building.''}} &
{\scriptsize \parbox{0.18\linewidth}{\centering ``A parking sign stands on a sidewalk near a building with green awnings.''}} &
{\scriptsize \parbox{0.18\linewidth}{\centering ``A symmetrical geometric composition with intersecting planes and warm yellow tones.''}} \\
{\small (a)} & {\small (b)} & {\small (c)} & {\small (d)} & {\small (e)} \\
\end{tabular}
\caption{Qualitative comparison between BlendFusion path-traced renders (top) and SDXL-generated images (bottom) conditioned on the same captions (shown below each pair). The diffusion-generated images exhibit geometric hallucinations (a, c), semantic drift (b), illegible text (d), and structural incoherence (e).}
\label{fig:qualitative}
\end{figure*}

To further motivate the use of path-traced renders over diffusion-generated synthetic data, we present a qualitative comparison between BlendFusion images and images generated by Stable Diffusion XL (SDXL) 1.0~\cite{podell2023sdxl} using the same captions from our dataset. For each sample, we use the BlendFusion caption as the text prompt for SDXL and compare the resulting image against the original Blender render. Several categories of artifacts are apparent in the diffusion-generated images. Figure~\ref{fig:qualitative} shows representative pairs.

\paragraph{Geometric hallucination.}
SDXL frequently hallucinates geometric structure absent from the described scene: in (a) a single wall-mounted cabinet becomes a floor-to-ceiling grid, and in (c) a simple post becomes an oversized concrete monolith that is not geometrically coherent.

\paragraph{Semantic drift.}
Rather than faithfully reconstructing specific scenes, SDXL tends to generate ``typical'' instances of the described concept. In (b), an oblique street view of a green building becomes a generic front-facing caf\'{e} facade with saturated colors, where the outdoor chairs and tables are physically impossible, with legs merging into one another.

\paragraph{Text rendering failures.}
Diffusion models struggle with legible text due to character-blind text encoders: in (d), the Blender render shows a clear ``P'' parking sign and a lamp post, whereas the SDXL output omits the parking sign entirely, renders the lamp post as a garbled artifact, and produces illegible text on storefronts. Training on such data risks teaching downstream models that scrambled glyphs are valid representations of text.

\paragraph{Structural incoherence.}
In (e), a geometric corridor with converging walls is rendered by SDXL as a chaotic structure with no physically consistent interpretation. Such failures are a known limitation of diffusion models when prompts require compositional spatial reasoning~\cite{podell2023sdxl}.

\paragraph{Implications for training data.}
These artifacts are not merely cosmetic. When diffusion-generated images are used as training data for other generative models, the hallucinated structures, semantic biases, and geometric inconsistencies propagate into the learned distribution, leading to Model Autophagy Disorder (MAD) \cite{alemohammad2023selfconsuminggenerativemodelsmad}, where quality and diversity progressively degrade across generations. Recent work further demonstrates that model collapse exhibits distinct and compounding failure modes in multi-modal systems combining vision and language models~\cite{fan2025multimodal}. By grounding our dataset in physically rendered scenes, BlendFusion avoids injecting these artifacts into the training pipeline entirely, providing a complementary source of synthetic data that is geometrically faithful by construction.

\subsection{Ablation: Object-Centric Camera Placement}

We evaluate the effect of our object-centric camera placement strategy by comparing it against two object-agnostic camera sampling approaches. For each method, 5,000 candidate frames per scene is rendered and compared against the BlendFusion dataset. 

\textbf{Random viewpoint sampling.}
This baseline samples camera positions uniformly within the scene's axis-aligned bounding box. Each sampled position produces one image by choosing a random azimuth direction and a random elevation within $\pm30^\circ$.

\textbf{Anchor-based camera sweep.}
This baseline samples anchor points within the scene bounding box and renders eight outward-facing views per anchor by sweeping azimuth angles every $45^\circ$ while keeping elevation fixed at $0^\circ$. Unlike our method, the camera does not explicitly target scene objects.

For both baselines we evaluate two spatial sampling variants: \textit{uniform sampling} (where every point is uniformly sampled from within the bounding box) and \textit{grid sampling}, where the scene's bounding box is divided into $n^3$ equal grids and equal number of points are picked uniformly from each grid to encourage spatial coverage. We pick $n=4$ to ensure that the grids are neither too small nor too large.

To evaluate robustness across scenes with different difficulty levels, we select three environments with \textbf{high}, \textbf{medium}, and \textbf{low} filtering rates under our pipeline: Emerald Square (high), Sun Temple (medium), and Barbershop (low).

Table~\ref{tab:camera_ablation} reports the percentage of rendered frames passing heuristic filtering and VLM filtering. Across all scenes, the object-centric strategy consistently produces the highest number of usable samples. In particular, the improvement is most pronounced in difficult scenes such as Barbershop, where naive camera sampling frequently produces empty or poorly framed views. These results demonstrate that explicitly targeting objects significantly improves the efficiency of synthetic dataset generation.

\begin{table}[t]
\centering
\footnotesize
\setlength{\tabcolsep}{4pt}
\begin{tabular}{l l cc}
\hline
Scene & Method & Heuristic (\%) & VLM (\%) \\
\hline
Barbershop & Object-centric & \textbf{16.1} & \textbf{5.2} \\
Emerald Square & Object-centric & \textbf{62.0} & \textbf{53.4} \\
Sun Temple & Object-centric & 23.8 & \textbf{15.2} \\
\hline
\end{tabular}

\vspace{4pt}

\begin{tabular}{l l cc cc}
\hline
Scene & Method & \multicolumn{2}{c}{Heuristic (\%)} & \multicolumn{2}{c}{VLM (\%)} \\
 &  & Uniform & Grid & Uniform & Grid \\
\hline
\multirow{2}{*}{Barbershop}
& Random view  & 11.6 & 10.5 & 1.3 & 1.4 \\
& Anchor sweep & 10.4 & 10.7 & 0.6 & 1.3 \\

\multirow{2}{*}{Emerald Square}
& Random view  & 43.8 & 43.6 & 24.1 & 23.6 \\
& Anchor sweep & 46.0 & 46.5 & 19.3 & 19.2 \\

\multirow{2}{*}{Sun Temple}
& Random view  & 31.6 & 31.9 & 3.1 & 3.3 \\
& Anchor sweep & 36.9 & \textbf{38.1} & 5.7 & 5.4 \\
\hline
\end{tabular}
\caption{Ablation of camera placement strategies. Our object-centric sampling consistently produces more usable samples after filtering compared to object-agnostic camera sampling baselines.}
\label{tab:camera_ablation}
\end{table}
\section{Conclusion and Limitations}

We presented \textbf{BlendFusion}, an automated pipeline for generating large-scale image--caption datasets from 3D scenes using physically-based rendering. Our approach combines object-centric camera placement, heuristic and VLM-based filtering, automatic caption generation, and diversity-aware sampling to produce high-quality image--caption pairs. Using this framework, we constructed \textbf{FineBLEND}, a dataset of 7,500 rendered image--caption pairs. Quantitative analysis shows that FineBLEND achieves CLIPScore distributions comparable to widely used datasets such as MS-COCO and Conceptual Captions, indicating strong semantic alignment between rendered images and generated captions. These results suggest that physically rendered synthetic data can serve as a scalable and controllable alternative to web-scraped training datasets.

The current FineBLEND dataset contains 7,500 image–caption pairs, which is smaller than the large-scale datasets typically used for training modern diffusion models. Due to computational constraints associated with large-scale path tracing and VLM-based filtering, we limit the dataset size in this study. Our goal in this work is to demonstrate the effectiveness of the BlendFusion pipeline with the hope that future work can scale this pipeline to much larger collections of 3D scenes to produce substantially larger datasets.

We also do not train a diffusion model directly on FineBLEND in this study. Nevertheless, the dataset exhibits strong image--text alignment and physically consistent geometry, suggesting it is well-suited for generative model training. A full empirical evaluation of models trained on FineBLEND is left to future work.

Lastly, the rendered images exhibit lower photorealism compared to natural photographs. One possible direction is diffusion-based refinement or ControlNet-style \cite{zhang2023adding} post-processing to enhance visual realism while preserving the physically grounded structure of the renders. At the same time, the domain differences between rendered and real images may provide useful diversity during training, as prior work has shown that synthetic datasets can improve generalization when combined with real-world data \cite{richter2016playingdatagroundtruth, 7780721}. 
{
    \small
    \bibliographystyle{ieeenat_fullname}
    \bibliography{main}
}

\appendix
\onecolumn

\section{Prompts}
\label{app:prompts}

\begin{tcolorbox}[colback=blue!5,colframe=blue!50!black,title=VLM Filtering Prompt]
\begin{verbatim}
You are evaluating a low-resolution synthetic render for a captioning dataset.

Decide if the image is GOOD or BAD.

Output format (exactly one line):

- "GOOD: <brief factual reason>"
- "BAD: <brief factual reason>"

Goal: keep only images that are EASY to caption accurately (specific nouns 
+ attributes).

Reject images that would force a vague caption (e.g., "a close-up of something").

PASS CONDITIONS (either is enough)
A) Object-centric: a recognizable object/character is shown with enough context
to name it.
B) Scene-centric: a recognizable scene/environment is shown (forest, room, 
street, landscape).

HARD REJECT (always BAD)
1) EXTREME CROP / CLOSE-UP
- BAD if the view is an extreme close-up or partial fragment such that 
  the subject cannot be confidently named.
- BAD if the frame is dominated by a single surface/part (e.g., cheek, wall, 
  texture) without context.
- BAD if >30% of the subject is cut off OR the crop removes key identifying 
  parts (e.g., head missing, face half missing, object mostly out of frame).
2) IDENTIFIABILITY FAILURE
- BAD if you cannot identify WHAT it is (object type OR scene type) in one 
  short noun phrase.
3) RENDER / SYNTHETIC ERRORS
- BAD if obvious rendering artifacts exist: clipping/interpenetration,
  broken geometry, missing textures/materials,
  NaN/black patches, fireflies/bright speckles, extreme distortion.
4) VISIBILITY FAILURE
- BAD if too dark/bright/blurred/noisy to recognize major shapes and boundaries.
- BAD if mostly blank/black/solid color.

FRAMING RULES
- Object-centric GOOD only if the full object OR a clearly intentional, 
  informative partial view is shown.
  (Example acceptable partial: "close-up of a clock face" where it is clearly 
    a clock.)
- Scene-centric GOOD only if the scene layout is readable and not dominated by 
  an uninformative foreground occluder.
OCCLUSION
- GOOD if occlusion is natural and still captionable.
- BAD if key content is blocked or inexplicably obscured.
INSTRUCTIONS
- Low resolution alone is NOT a reason to reject.
- Be strict: if the best caption would be vague, mark BAD.
- Keep the reason short and factual (5–10 words).
- Output exactly one line.
\end{verbatim}

\end{tcolorbox}

\begin{tcolorbox}[colback=red!5,colframe=red!50!black,title=VLM Filtering Prompt]
\begin{verbatim}
Describe the image with a single factual caption.

Output format:
- A single sentence caption (no extra text).

Rules:
- Describe only what is clearly visible in the image.
- Do NOT guess or hallucinate unseen objects.
- Do NOT include opinions, emotions, or storytelling.
- Use concrete nouns and simple attributes (colors, materials, positions).
- Mention the main objects and the environment if visible.
- If it is a scene (e.g., landscape, forest, room), describe the scene 
  structure.
- If it is an object, mention the object and its context/background.

Style guidelines:
- 8–20 words.
- Neutral, factual tone.
- No phrases like "this image shows" or "there is".
- Avoid unnecessary adjectives.

Examples:

GOOD:
"Three trees standing on a grassy field under a cloudy sky."
"A wooden chair placed next to a small table in a bright room."
"A red car parked on a paved road beside a row of buildings."

BAD:
"This image shows a beautiful scene with some trees."
"A very nice and detailed picture of a forest landscape."
"Possibly a statue or object that looks like metal."

Return only the caption.
\end{verbatim}
\end{tcolorbox}


\end{document}